\documentclass[letterpaper, 10 pt, journal, twoside]{ieeetran}

\IEEEoverridecommandlockouts 

\usepackage{algorithmic}
\usepackage{amsmath, amssymb, amsfonts}
\usepackage{graphics} 
\usepackage{graphicx} %
\usepackage{float} %
\usepackage{subfigure} %
\usepackage{threeparttable}
\usepackage{multirow}
\usepackage{array}
\usepackage{algorithm}
\usepackage{algorithmic}
\usepackage[normalem]{ulem}
\usepackage[colorlinks,linkcolor=red]{hyperref}
\usepackage{cite}
\usepackage{amsmath}

\newcommand{\norm}[1]{\left\lVert#1\right\rVert}
\newcommand{\nodata}[0]{{\enspace\enspace-\enspace\enspace}}

\begin{document}


\title{Efficient Heatmap-Guided 6-Dof Grasp Detection in Cluttered Scenes}

\author{Siang Chen$^{1,2,4,\dagger}$, Wei Tang$^{3,4,\dagger}$, Pengwei Xie$^{1,4}$, Wenming Yang$^{3,4}$, Guijin Wang$^{1,2,4,*}$%
    \thanks{$^{1}$Department of Electronic Engineering, Tsinghua University, Beijing 100084, China.}
    \thanks{$^{2}$Shanghai AI Laboratory, Shanghai 200232, China.}
    \thanks{$^{3}$Department of Electronic Engineering, Shenzhen International Graduate School, Tsinghua University, Shenzhen 518071, China.}
    \thanks{$^{4}$Beijing National Research Center for Information Science and Technology (BNRist), China.}
    \thanks{$^{\dagger}$These authors contributed equally.}
    \thanks{*Correspondence: {\tt wangguijin@tsinghua.edu.cn.}}
}

\maketitle

\begin{abstract}

Fast and robust object grasping in clutter is a crucial component of robotics. Most current works resort to the whole observed point cloud for 6-Dof grasp generation, ignoring the guidance information excavated from global semantics, thus limiting high-quality grasp generation and real-time performance. In this work, we show that the widely used heatmaps are underestimated in the efficiency of 6-Dof grasp generation. Therefore, we propose an effective local grasp generator combined with grasp heatmaps as guidance, which infers in a global-to-local semantic-to-point way. Specifically, Gaussian encoding and the grid-based strategy are applied to predict grasp heatmaps as guidance to aggregate local points into graspable regions and provide global semantic information. Further, a novel non-uniform anchor sampling mechanism is designed to improve grasp accuracy and diversity. Benefiting from the high-efficiency encoding in the image space and focusing on points in local graspable regions, our framework can perform high-quality grasp detection in real-time and achieve state-of-the-art results. In addition, real robot experiments demonstrate the effectiveness of our method with a success rate of 94\% and a clutter completion rate of 100\%. Our code is available at \href{https://github.com/THU-VCLab/HGGD}{https://github.com/THU-VCLab/HGGD}.
\end{abstract}


\section{INTRODUCTION}

\IEEEPARstart{O}{bject} grasping is a critical component of robotics in manufacturing, service, medical assistance, etc. Despite its vital importance, fast and accurate grasping is still challenging for robots. Recent advances in deep learning have enabled data-driven methods to generalize to unseen objects. Representative methods \cite{morrison2018closing,kumra2020antipodal,yu2022se} generate grasp configurations as oriented grasp rectangles by adopting pixel-wise heatmaps to represent planar grasps, achieving good performance in simple scenarios with high-efficiency. However, such a representation forces the gripper perpendicular to the camera plane, limiting the applications.

\begin{figure}[ht]
    \centering
    \includegraphics[width=8cm]{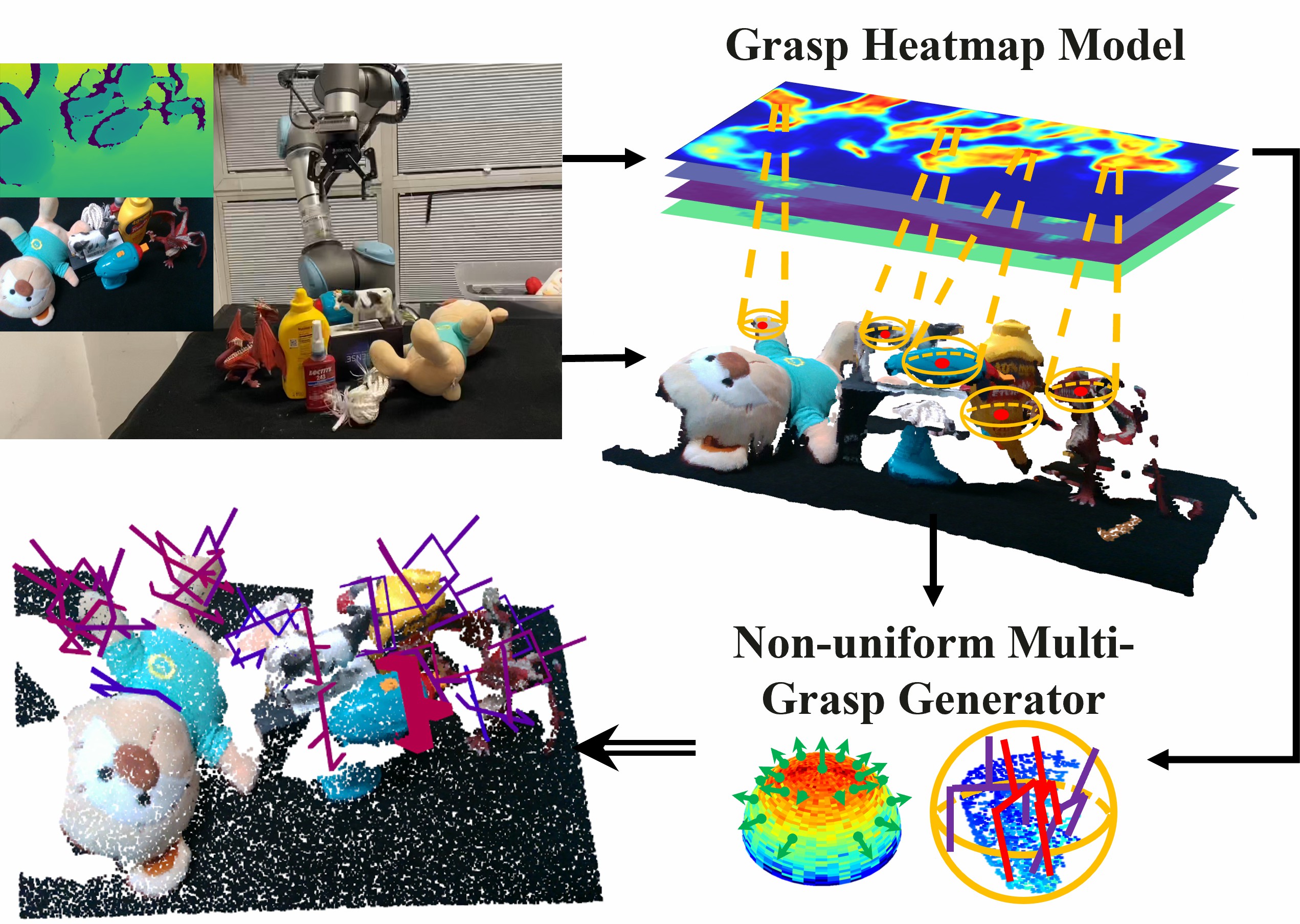}
    \caption{The key insight of our method is generating the grasp heatmaps as guidance for regional geometric feature mining and further grasp pose generation via a novel local grasp generator.}
    \label{fig:teaser}
\end{figure}

Recently, considerable attention has been paid to 6-Dof grasping, enabling robots to grasp from arbitrary directions. \cite{ten2017grasp,liang2019pointnetgpd} adopt a sampling-evaluation strategy which is extremely time-consuming. \cite{ni2020pointnet++,qin2020s4g} directly regress grasp attributes from the extracted per-point features. However, these methods tend to generate unreliable grasps since ignoring the local contextual geometric information. To ameliorate this issue, recent methods \cite{zhao2021regnet,wei2021gpr} encode locally aggregated features to generate or fine-tune grasp poses. Despite the impressive performances, these approaches still struggle to perform high-quality 6-Dof grasp detection in real-time. 

Accordingly, motivated by the satisfying performance brought by the widely used heatmaps in object detection \cite{zhou2019objects}, human pose estimation \cite{osokin2018real} and planar grasping \cite{morrison2018closing}, this paper extends heatmaps for high-quality 6-Dof grasp generation with high-efficiency through a carefully designed local grasp generator. We propose a novel efficient 6-Dof grasp detection framework in cluttered scenes. As shown in Fig. \ref{fig:teaser}, our key insight is constructing the grasp heatmaps as guidance to aggregate local points into graspable regions for further grasp pose generation. Thus, it can distinguish the graspable regions and significantly reduce the input size. Specifically, Gaussian encoding and the grid-based strategy are introduced to predict grasp confidence and attribute heatmaps robustly and efficiently. Furthermore, we design a local grasp generator combined with a novel non-uniform anchor sampling mechanism to precisely estimate spatial rotations in local aggregated regions, enabling anchor sampling with minimal fitting error between anchors and the actual grasp rotation distributions. Notably, our pipeline can fully leverage the global semantic and local geometrical representations benefiting from our semantic-to-point feature fusion.

In summary, our primary contributions are as follows: 
\begin{itemize}
    \item We propose a novel global-to-local semantic-to-point 6-Dof grasp detection framework, achieving state-of-the-art performance in real-time through a low-cost training procedure.
    \item Grasp attribute heatmaps are predicted by the proposed Gaussian encoding and the grid-based strategy, significantly improving the encoding efficiency and reducing the local input size for grasp generation.
    \item A local grasp generator with a novel non-uniform anchor sampling mechanism is designed to generate dense grasps precisely, and an extra local semantic-to-point feature fusion makes grasp generation more robust.
\end{itemize}

\section{RELATED WORKS}

Existing grasp detection methods can be roughly divided into model-based and model-free. The model-based methods \cite{xiang2017posecnn,zeng2017multi,wang2019densefusion,he2020pvn3d} transfer the task to object pose estimation and project grasps from the pre-prepared database. In contrast, model-free methods like \cite{lenz2015deep,morrison2018closing,wang2019efficient} consider the problem as grasp rectangle detection in images, still limited in some scenarios since the grippers can only be perpendicular to the camera plane. 

Recently, 6-Dof grasping has been widely researched due to the flexibility that the gripper can approach the object from arbitrary directions. \cite{ten2017grasp,liang2019pointnetgpd} propose the sample-evaluation strategy to select high-quality grasps from a large number of grasp proposals. \cite{qin2020s4g} directly regresses grasps from per-point features extracted by PointNet++. Different from these methods, our method designs a global-to-local semantic-to-point framework to detect grasps, avoiding time-consuming grasp candidates sampling and significantly increasing the quality of generated grasps. 

Based on the large-scale grasp dataset and benchmark \cite{fang2020graspnet}, \cite{gou2021rgb} utilize RGB images to generate pixel-wise orientation heatmaps and filters unreasonable grasps with point clouds. Similarly, \cite{wang2021graspness} proposes point-wise graspness to represent the possibility of grasp locations and approach directions. Unlike these methods, we predict heatmaps as guidance to aggregate local points into graspable regions and only focus on these regions for further grasp pose generation.

To fully exploit the local contextual features, \cite{wei2021gpr} refines grasp poses locally, and \cite{zhao2021regnet} presents a network based on grasp regions and gripper closing areas to obtain local shape information of grasps. More targeted than them, our method constructs heatmaps to distinguish graspable regions and perform grasp orientation prediction and center refinement on these regions, fully leveraging the fusion of semantic and geometrical representations and facilitating real-time detection performance. Besides, to precisely predict the spatial rotation angle or the approach view, \cite{zhao2021regnet,wei2021gpr,fang2020graspnet} introduce the anchor-based strategy, achieving higher rotation prediction accuracy \cite{zou2023object} than directly regressing. Unlike sampling pre-defined anchors from the surface of a unit sphere or using Fibonacci lattices \cite{gonzalez2010measurement}, we design a novel non-uniform anchor sampling mechanism to predict the spatial rotation.

\begin{figure}[ht]
\centering
    {\includegraphics[width=7cm]{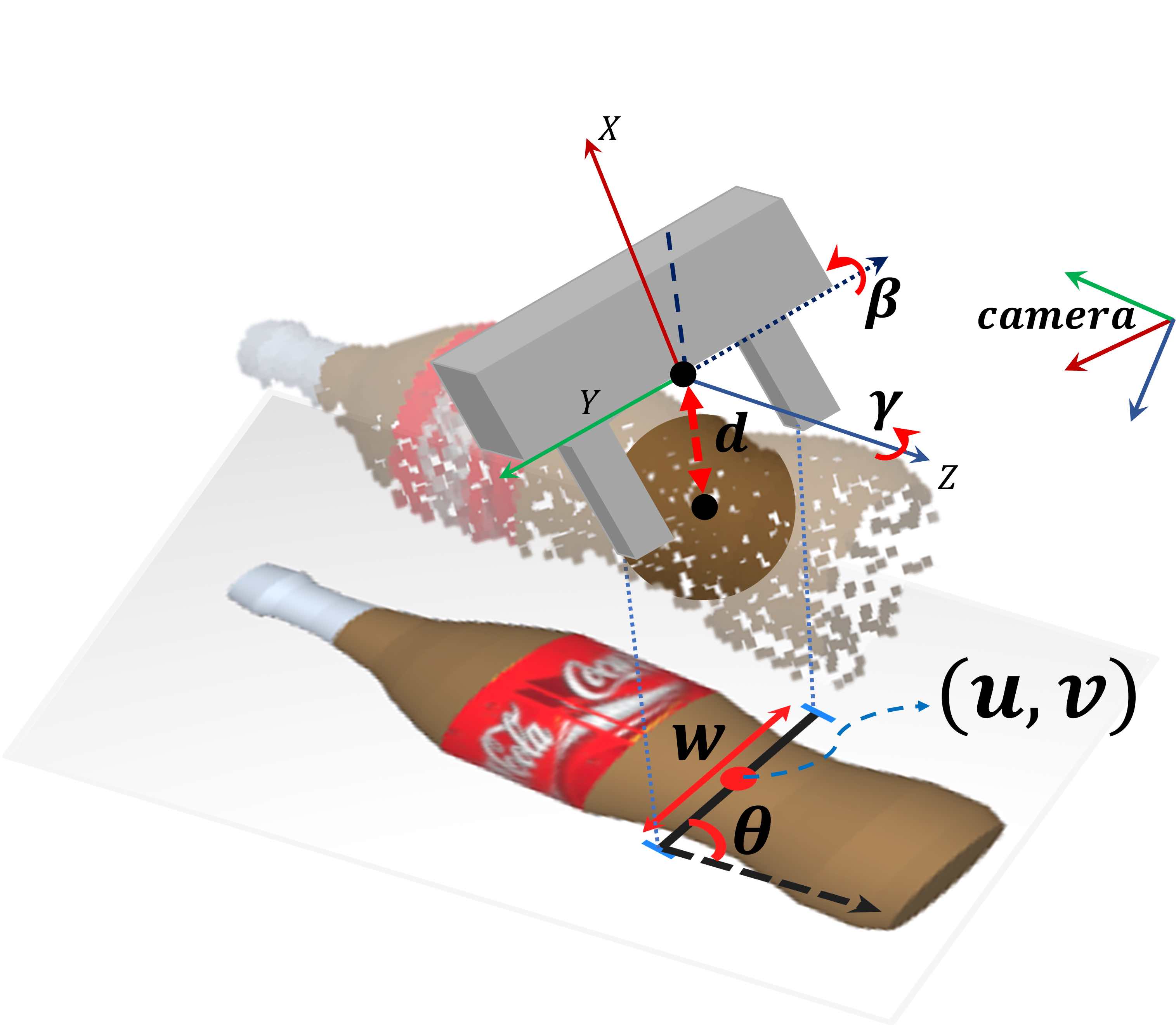}}
        \caption{Proposed grasp representation as $(u,v,\theta,w,d,\gamma,\beta)$.}
    \label{fig:representation} 
\end{figure}

\begin{figure*}[ht]
\centering
    {\includegraphics[width=17cm]{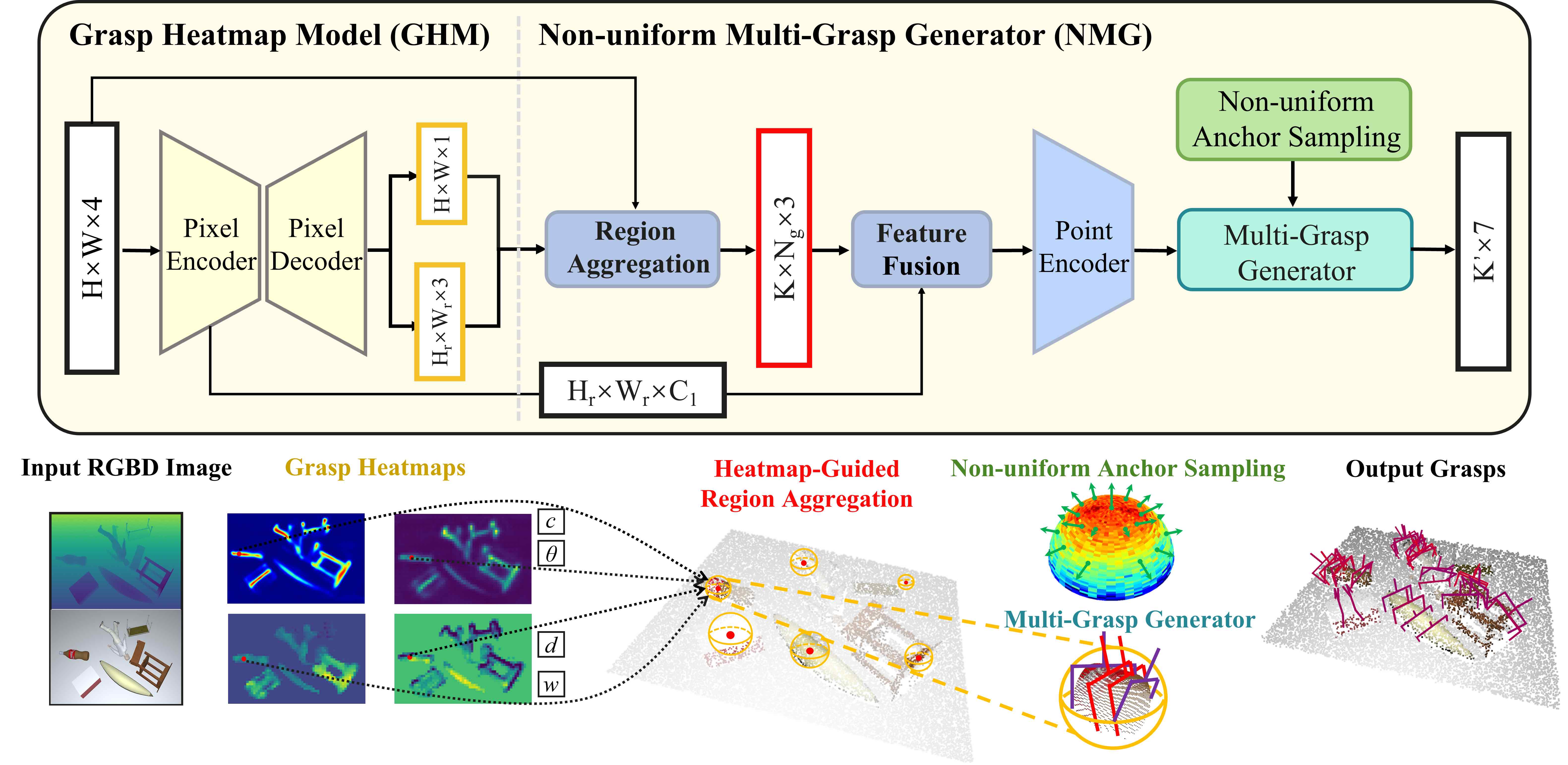}}
        \caption{The architecture of HGGD. Taking a monocular RGBD image as input, GHM generates grasp confidence heatmap $Q_c$ and grided attributes heatmaps $(Q_{\theta},Q_w,Q_d)$. Then NMG transfers the depth image to the point cloud through camera intrinsics $\mathbf{c}$ for region aggregation under the guidance of heatmaps. Feature fusion and the point encoder extract regional features fused with semantic information from GHM. Finally, a multi-grasp generator combined with a novel non-uniform anchor sampling mechanism utilizes the fusion features to output the grasps.}
    \label{fig:framework} 
\end{figure*}

\section{Problem Statement}

Given a monocular RGBD image $\boldsymbol{\chi} \in R^{ H \times W \times 4}$ and corresponding camera intrinsics $\mathbf{c}$, our framework aims to efficiently learn parallel-jaw grasp configurations $\mathbf{G}$ in a cluttered scene. Equivalent to former definitions of the grasp in \cite{liang2019pointnetgpd,wei2021gpr,fang2020graspnet}, to better fit proposed global-to-local semantic-to-point framework, one grasp pose $\boldsymbol{g}$ is defined as: 
\begin{equation*}
    \boldsymbol{g} = (u, v, \theta, w, d, \gamma, \beta).
\end{equation*}

As Shown in Fig. \ref{fig:representation}, rather than directly locating grasp centers in the 3-dimensional coordinate, we adopt a 2-dimensional tuple $(u, v)$ to represent the grasp center in the image plane. $(u, v, \theta, w)$ with another fixed grasp height parameter $h$ compose an oriented rectangle, representing a 4-Dof grasp similar to \cite{jiang2011efficient}. $\gamma, \beta\in [-\frac{\pi}{2}, \frac{\pi}{2}]$ are grasp Euler angles among axis-z and axis-y in the gripper coordinate, while $d$ denotes the depth offset from the grasp center to corresponding point cloud surface point. Refer to appendices for more details of our data preparation.

\section{METHOD}

\subsection{Overview}

We aim to efficiently generate high-quality and abundant grasps $\mathbf{G}$ in a novel global-to-local semantic-to-point way, with monocular RGBD images and camera intrinsics $\mathbf{c}$. As illustrated in Fig. \ref{fig:framework}, instead of directly processing the observed point cloud, our method inputs RGBD images to encode grasp heatmaps as graspable region guidance efficiently. With the critical heatmap guidance, only these regions' semantic and geometrical representations are extracted and fused. Then a novel local grasp generator enables \textbf{HGGD} (\textbf{Heatmap-Guided 6-Dof Grasp Detection}) to detect grasps with high-quality and diversity in real-time. Our model comprises two sub-modules: \textbf{Grasp Heatmap Model} (\textbf{GHM}) and \textbf{Non-uniform Multi-Grasp Generator} (\textbf{NMG}). 

\textbf{GHM} preprocesses input RGBD images, extracts semantic features with an efficient CNN, and further generates {four grasp heatmaps as guidance}. Gaussian encoding is applied to encode the grasp ground truths, assisting in locating graspable areas more precisely. The grid-based strategy transforms the discontinuous pixel-wise regression into the prediction based on neighborhood similarity, which makes heatmap generation more robust for each local region.

\textbf{NMG} utilizes the heatmaps generated in GHM as guidance to aggregate local points into graspable regions and detects grasps through a light-weighted point encoder in each region. The proposed non-uniform anchor sampling mechanism adopted in NMG improves grasp quality by fitting the ground truth distribution better. Furthermore, a novel semantic-to-point feature fusion module is applied to detect grasps more robustly. 

\subsection{Grasp Heatmap Model}

\begin{figure}
\centering
    {\includegraphics[width=8.5cm]{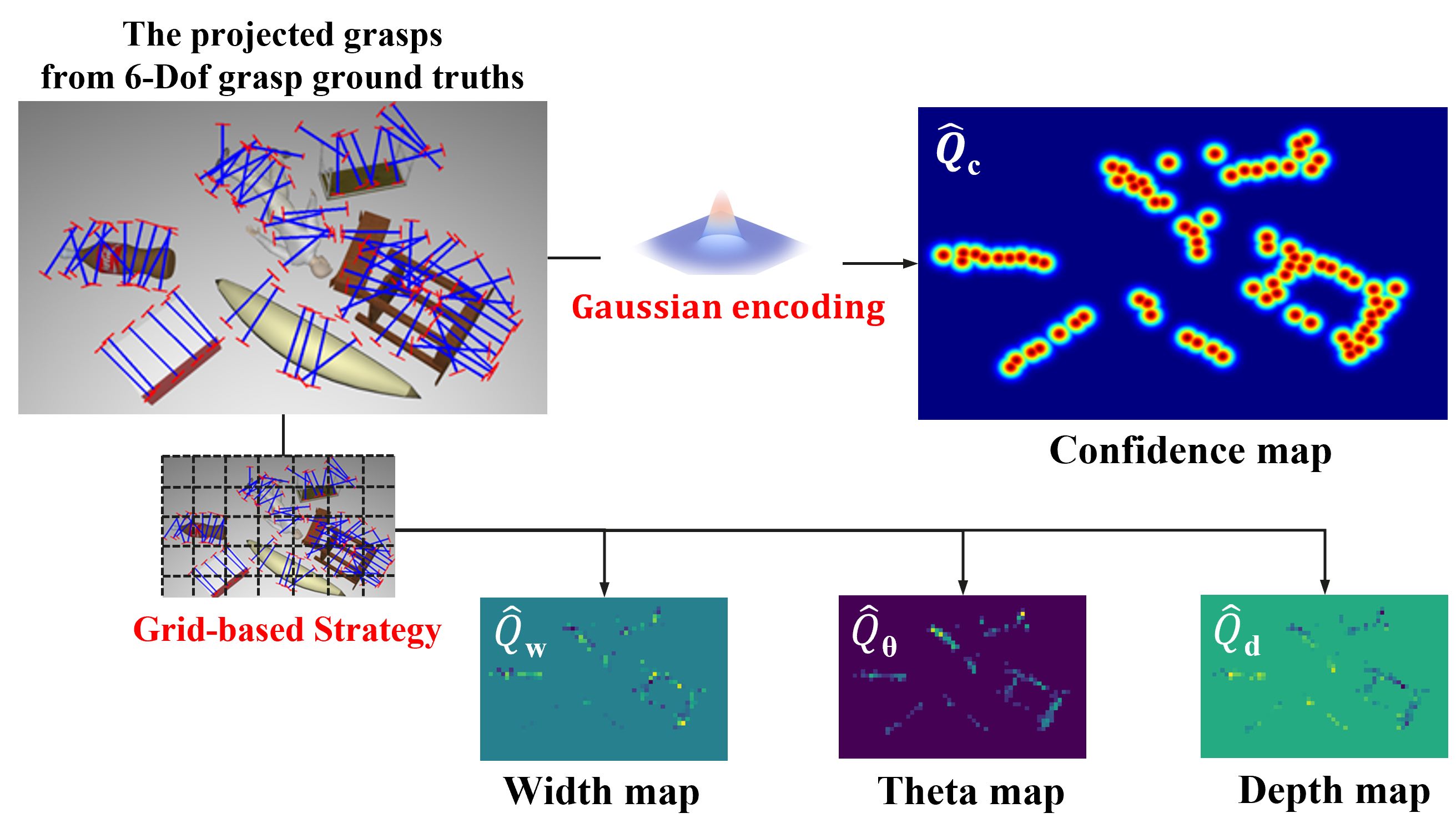}}
        \caption{Visualization of how the ground truth 6-Dof grasps are projected. Grasp confidence heatmap ${\hat{Q}_{c}}$ and attribute heatmaps $(\hat{Q}_{\theta},\hat{Q}_{w},\hat{Q}_{d})$ are encoded with Gaussian kernel and grids, respectively.}
    \label{fig:extraction} 
\vspace{-0.5cm} 
\end{figure}

GHM is an encoder-decoder model containing two output branches, the confidence branch aiming at constructing grasp confidence heatmap $Q_{c}$ and the attribute branch aiming at generating attribute heatmaps $(Q_{\theta},Q_{w},Q_{d})$. Inspired by \cite{duan2019centernet, zhou2018fully}, we apply Gaussian encoding and the grid-based strategy to decouple this task on account of different peculiarities between heatmaps. As is shown in Fig. \ref{fig:extraction}, the ground truth 6-Dof grasps are projected to the image plane and encoded to heatmaps $({\hat{Q}_{c}},\hat{Q}_{\theta},\hat{Q}_{w},\hat{Q}_{d})$.


The Gaussian encoding strategy adopts a 2D Gaussian kernel to encode the projected grasp ground truth centers before training. This approach effectively highlights grasp centers without ignoring nearby pixels, because pixels nearby will also serve as helpful guidance for further grasp detection. The value of pixel $(u,v)$ in the confidence heatmap for training can be calculated by
\begin{equation}
q=\exp\left(-\frac{\left(u-u_0\right)^{2}+\left(v-v_0\right)^{2}}{2 \sigma_{g}^{2}}\right),
\end{equation}
where $(u_0,v_0)$ denotes the center point of a grasp ground truth, and $\sigma_g$ is the standard deviation which depends on the width of each grasp. Supervised by $\hat{Q}_{c}$, the confidence branch applies pixel-wise classification to predict the $Q_c$.

The proposed grid-based strategy encodes and predicts grasp attributes $(\theta,w,d)$ within a specific local grid instead of direct pixel-wise regression. Grasp attributes usually have high similarity in these areas due to the similar geometric structure. Thus, by fully exploiting the similarity of adjacent grasps, more robust grasp attribute prediction can be accomplished. Concretely, the full-scale image is divided into $H_r \times W_r$ grid cells with side length $r$. Based on the oriented anchor box mechanism in \cite{zhou2018fully,depierre2021scoring}, for each grid cell, $k_a$ multiple oriented anchors are introduced with uniformly sampled angles. Therefore, the ground truth $\theta$ can be assigned to the nearest anchor. We obtain the number distribution of anchors in each grid, and a sigmoid function is applied afterward to acquire  $\hat{Q}_{\theta}$. Besides, we calculate the average normalized $w, d$ in grids to obtain ground truth attribute heatmaps $(\hat{Q}_w, \hat{Q}_d)$. Supervised by these heatmaps, in a patch-wise manner, the attribute branch predicts $Q_{\theta}$ through the combination of anchor classification and offset regression, and estimates $(Q_w, Q_d)$ via direct regression.

Previous methods like \cite{morrison2018closing, kumra2020antipodal,gou2021rgb} encode grasps as pixel-wise rectangles, suffering from two defects. First, they fail to highlight the importance of the most considerable grasping probability at the center point \cite{yu2022se}. Second, ground truth attribute heatmaps $(\hat{Q}_{\theta},\hat{Q}_w,\hat{Q}_d)$ are not smooth as the confidence heatmap $\hat{Q}_c$ due to the relatively dense grasp annotations in cluttered scenes. In contrast, the designed GHM can highlight grasp centers and predict more robust grasp attributes, especially in cluttered scenes. 

\subsection{Non-uniform Multi-Grasp Generator}

NMG takes the heatmaps and the scene point cloud as input and aggregates multiple graspable local areas efficiently under the guidance of the heatmaps. Subsequently, utilizing the grasp attributes in each grid, NMG predicts the remaining grasp rotation attributes and refines the former generated ones by the local features to generate multiple grasps. The proposed non-uniform anchor sampling mechanism improves the grasp quality and the novel semantic-to-point feature fusion contributes to the robustness of the detected grasps. According to different functions, the overall structure of NMG can be split into two parts, Heatmap-Guided Region Aggregation, and Non-uniform Multi-Grasp Generator.

\subsubsection{\textbf{Heatmap-Guided Region Aggregation}}

The first part of NMG processes heatmaps and the point cloud into useful local features, consisting of two steps: the region aggregation and the feature fusion.


Firstly, the region aggregation
aggregates local points into graspable regions under the guidance of heatmaps from GHM for the subsequent multi-grasp generator. Concretely, the grasp confidence heatmap is downsampled with bilinear interpolation to $H_r \times W_r$, which is the same shape as the attribute heatmaps. Then top $k_{center}$ grids with the highest predicted confidence are selected, containing $k_{center}$ local peaks in total as regional centers. This grid-based selection suppresses center density to reduce the aggregation of duplicate areas. During training, $k_{center}$ is set to a larger number to ensure most graspable local regions are extracted. During inference, it is convenient to adjust $k_{center}$ to achieve grasp detection with different coverage rates.

Afterward, we transform the pixel centers $(u,v)$ with corresponding depth $d$ to point centers $(x,y,z)$ in 3D space using camera intrinsics $\mathbf{c}$. As Fig. \ref{fig:framework} suggests, a ball query \cite{qi2017pointnet} is utilized to crop the points within a sphere with a radius of the predicted grasp width $w$ for each center. In each local ball region, $N_g$ points are sampled by farthest point sampling to reduce the complexity of subsequent calculations while maintaining the local geometric information to the maximum extent.

\begin{figure}[ht]
    \centering
    \includegraphics[width=8.5cm]{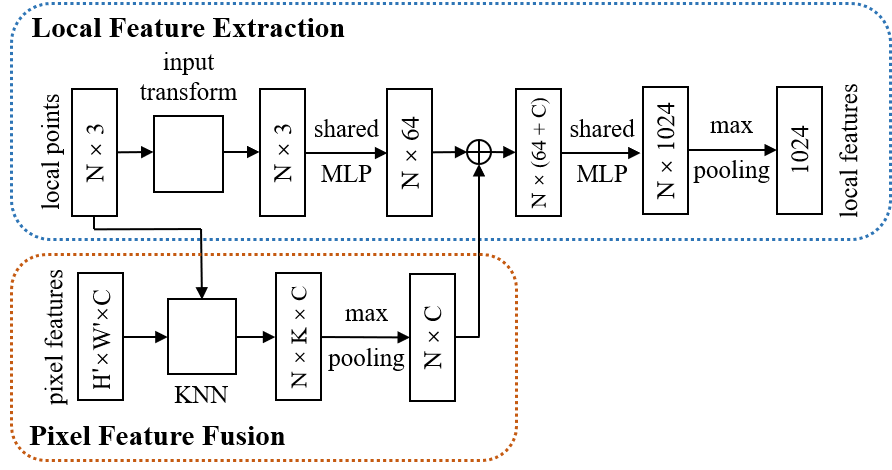}
    \caption{The pipeline of local region feature extraction
with semantic-to-point feature fusion.}
    \label{fig:fusion}
\end{figure}


Additionally, pixel-wise features extracted in GHM contain rich semantic information, which can reasonably supplement local point clouds. Therefore, we design a new light-weighted PointNet-based \cite{qi2017pointnet} network with semantic-to-point feature fusion for local feature extraction. The overall process is illustrated above in Fig. \ref{fig:fusion}, where 
pixel features are grouped via a KNN operation to each local point. Then we integrate the pooled features with point features by point-wise concatenation for further feature extraction. By conducting KNN grouping and combining shared MLP (Multi-Layer Perceptrons) with max-pooling, we consider both local geometric and semantic information in the following grasp generator, increasing the grasp robustness, especially when point cloud input is unreliable.

\subsubsection{\textbf{Non-uniform Multi-Grasp Generator}}

\begin{figure}[t]
    \centering
    \includegraphics[width=8.5cm]{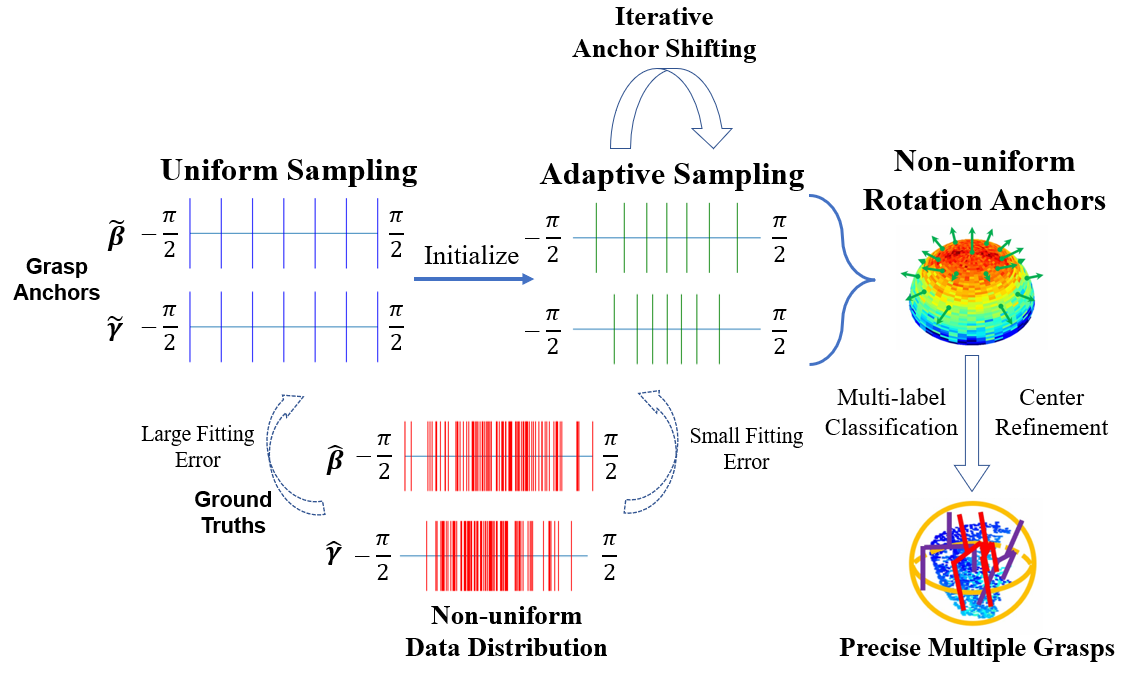}
    \caption{Visual illustration for the procedure of the anchor shifting algorithm and the multi-grasp generation.}
    \label{fig:nmg}
\end{figure}

The second part of NMG adopts anchor-based methods to detect grasps by the local features, consisting of the non-uniform anchor sampling and the multi-grasp generator. As Fig. \ref{fig:nmg} suggests, we generate non-uniform rotation anchors via iterative anchor shifting and conduct multi-label classification in each local region to generate multiple grasps. Notably, center refinement is applied to alleviate the errors resulted from the first stage and acquire more precise results.

With the local regions aggregated, only the 2D rotation $(\gamma,\beta)$ has to be determined. Because both angles are continuous in $[-\frac{\pi}{2},\frac{\pi}{2}]$, anchor-based methods prove to achieve better localization accuracy than direct regression \cite{zou2023object}. Most former works predefine approach vectors uniformly on a sphere surface for grasp rotation prediction \cite{zhao2021regnet,fang2020graspnet}, while the actual distribution proves to be uneven. Hence, there is an inevitable trade-off between rotation prediction accuracy and time efficiency, which means denser anchors provide better accuracy but slower speed. Unlike methods before, we present a novel anchor-shifting algorithm applied in the training process, gradually shifting our anchors to minimize the fitting error between anchors and the acquired grasp rotation distributions, successfully attaining higher performance with fewer anchors. To simplify the problem, we consider $\gamma$ and $\beta$ equally and take $\gamma$ as the example. The goal of the anchor shifting for the cumulative $K$ grasps and $k_{r}$ defined anchors ${\boldsymbol{\Tilde{\gamma}}}$ can be formulated below:
\begin{equation}
    \label{eq:anchor_error}
    \begin{gathered}
        {\boldsymbol{\Tilde{\gamma}}^*},\mathbf{B}_\gamma^* = \mathop{\arg\min}\limits_{{\boldsymbol{\Tilde{\gamma}}},\mathbf{B}_\gamma} \norm{\mathbf{B}_\gamma^T{\boldsymbol{\Tilde{\gamma}}}-{\boldsymbol{\hat{\gamma}}}}_2^2, \\
    \end{gathered}
\end{equation}
in which our grasp anchors are defined as ${\boldsymbol{\Tilde{\gamma}}}\in[-\frac{\pi}{2},\frac{\pi}{2}]^{k_r\times 1}$, and $\mathbf{B}_\gamma\in\{0,1\}^{k_r\times {K}}$ represents one-hot encodings of the nearest anchor's index for each grasp ground truth. ${\boldsymbol{\hat{\gamma}}}\in[-\frac{\pi}{2},\frac{\pi}{2}]^{{K}\times 1}$ is the rotation angles of the cumulative $K$ selected grasp ground truths during the training procedure. 

For dynamically and efficiently shifting our anchors during training, we solve the above problems via a block coordinate descent algorithm:

\begin{itemize}
    \item Get updated anchor encodings $\mathbf{B}^*_\gamma$ according to $\boldsymbol{\Tilde{\gamma}}$ by following simple comparison,
    \begin{equation}
        \label{eq:anchor_compare}
        \begin{gathered}
            \mathbf{B}_\gamma^{(i,j)}=
            \begin{cases}
                1 & \text{if}\ {\mathop{\arg\min}\limits_{k\in\{1,...,k_r\}}{\norm{\boldsymbol{\hat{\gamma}}^{(j)}-\boldsymbol{\Tilde{\gamma}}^{(k)}}}=i,} \\
                0 & \text{else.}
            \end{cases} \\
        \end{gathered}
    \end{equation}
    \item Fix $\mathbf{B}_\gamma$ and update $\boldsymbol{\Tilde{\gamma}}$: Eq.(\ref{eq:anchor_error}) reduces to a linear regression problem which can be solved with the least square method as
    \begin{equation}
        \label{eq:anchor_iter}
        {\boldsymbol{\Tilde{\gamma}}^*}={\left(\mathbf{B}_\gamma \mathbf{B}_\gamma^T\right)}^{-1}\mathbf{B}_\gamma{\boldsymbol{\hat{\gamma}}}.
    \end{equation}
\end{itemize}

We conduct anchor shifting during the whole training procedure, represented as the python-style pseudocode in Algorithm \ref{alg:anchor_shift}.

\begin{algorithm}[ht]
\caption{Non-uniform anchor shifting during training}
\label{alg:anchor_shift}
\textbf{Parameters:} $\boldsymbol{\Tilde{\gamma}},\boldsymbol{\Tilde{\beta}}\in[-\frac{\pi}{2},\frac{\pi}{2}]^{k_r\times1}$ - current anchors

$K$ - grasp number threshold, $T$ - shifting iterations

\textbf{Python-Style Pseudocode:} 
\begin{algorithmic}[1]
\STATE $Grasps=list()$ 
\WHILE{$training$}
\STATE {$G = GetGraspGroudTruthsInEachRegion()$}
\STATE $Grasps.extend(G)$
\IF{$len(Grasps)>K$}
\STATE $\boldsymbol{\hat{\gamma}},\boldsymbol{\hat{\beta}}=Grasps.\boldsymbol{\gamma},Grasps.\boldsymbol{\beta}$
\FOR{$t=1\rightarrow T$}
\STATE Get $\mathbf{B}_{\gamma,t},\mathbf{B}_{\beta,t}$ with $\boldsymbol{\Tilde{\gamma}}_{t-1},\boldsymbol{\Tilde{\beta}}_{t-1}$ per Eq.(\ref{eq:anchor_compare})
\STATE Update $\boldsymbol{\Tilde{\gamma}}_t,\boldsymbol{\Tilde{\beta}}_t$ with $\mathbf{B}_{\gamma,t},\mathbf{B}_{\beta,t}$ per Eq.(\ref{eq:anchor_iter})
\ENDFOR
\STATE $Grasps.clear()$ 
\ENDIF
\ENDWHILE
\end{algorithmic}
\label{alg:shift}
\end{algorithm}

At the same time with anchor shifting, supervised by local grasp ground truths, our multi-grasp generator takes the 
region aggregated features as input and combines anchors of the two angles to form a higher dimensional ($k_r^2$-class) multi-label classification problem. Then an MLP is utilized to generate the multi-label classification results and form multiple grasps in each local region. It is nonnegligible that the errors of the first stage, especially the errors of center localization, might affect the performance of the grasp generator. Thus, our multi-grasp generator not only predicts grasp rotation attributes but also refines grasp centers estimated in the first stage by regressing 3-dimensional center offsets for each anchor.

\subsection{Loss \& Implementation Details}

\subsubsection{\textbf{Loss}} The overall training objective of HGGD is the weighted sum of the heatmap losses and the anchor losses, formulated as:
\begin{equation*}
    L = \underbrace{{L_{Q_c}} + a \times L_{cls} + b \times L_{reg}}_{heatmap\ losses} + \underbrace{L_{anchor} + c \times L_{offset}}_{anchor\ losses},
\end{equation*}
where ${L_{Q_c}}$ represents the pixel-wise cross-entropy loss between the predicted grasp confidence $Q_{c}$ and the encoded ground truth $\hat{Q}_c$. We present a penalty-reduced focal loss \cite{lin2017focal} similar to \cite{zhou2019objects} to match the Gaussian-based heatmap. Another focal loss $L_{cls}$ is employed to supervise the multi-label classification \cite{zhang2013review} for $\theta$ learning. Moreover, a masked Smooth $L1$ loss $L_{reg}$ is adopted for all the regression problems in GHM. $L_{anchor}$ represents the local grasp rotation anchor classification loss calculated using focal loss, and $L_{offset}$ is a Smooth $L1$ loss adopted to predicted grasp center offsets of different rotation candidates.

\subsubsection{\textbf{Implementation Details}}

We adopt ResNet-34 \cite{he2016deep} as the backbone to build our pixel encoder for heatmap generation in GHM. The channel number is reduced to $128$ for better inference efficiency. With extra skip connections, the pixel decoder is composed of stacking transposed convolution layers to upsample the feature map and fit the output shape as the heatmaps. We set the input image resolution to $640\times 360$, the oriented anchor number $k_a$ to $6$, and the grid size $r$ to $8$. In NMG, during training, we set $k_{center}=128$ to cover areas as much as possible, while during the inference, $k_{center}$ is set to 32 and 48 for two datasets respectively. Anchor shifting iteration number $T$ is set to $1$ to get gentler anchor value movement. We aggregate local points with $N_g=512$ for each region and generate multiple grasps with $k_r=7$. End-to-end joint training is adopted. 



\section{DATASET \& EVALUATION METRICS}

\subsection{Dataset}

Grasp datasets can be roughly divided into real and synthetic according to the type of observations. GraspNet-1Billion \cite{fang2020graspnet} builds a large-scale grasp dataset in which the observations are captured in the real world. Although more than 1 billion grasp annotations are provided, it is still limited in the scale of objects, which are only 88 with similar shapes. 

On the other hand, simulating observations provides a more scalable alternative \cite{eppner2021acronym}. Thus, similar to \cite{qin2020s4g,zhao2021regnet}, we build a large-scale simulated grasp dataset based on ACRONYM \cite{eppner2021acronym}, containing mesh-based grasps generated with a physics simulator. We manually select 300 objects with various typical geometries from ShapeNetSem \cite{savva2015semantically} with supplementary textures for dataset construction. Then, we construct 500 cluttered scenes, each containing 6-8 objects, render 50 synthetic RGBD images from random perspectives for each scene, and project collision-free grasp labels to the camera frame. Additionally, we calculate each grasp a quality score to describe the force closure property following \cite{qin2020s4g}. We call this generated dataset TS-ACRONYM, which tells textured and scored. Besides, different from the \cite{fang2020graspnet,eppner2021acronym}, TS-ACRONYM contains compact scene-level grasp labels instead of dense object-level grasp labels, which avoids extra data transformation during training and reduces storage consumption. 

\begin{table}[t]
\caption{Results on TS-Acronym Dataset}
\begin{center}
\renewcommand\arraystretch{1.25}
\begin{threeparttable}
\begin{tabular}{c|c c c c}
\hline
\textbf{Method}&{\textbf{CR}} $\uparrow$ &{\textbf{CFR}} $\uparrow$ &{\textbf{AS}} $\uparrow$ &{\textbf{Time$^{1,2}$ (ms)}} $\downarrow$ \\
\hline
GPD (3 channels) \cite{ten2017grasp} & - & 69.3 \% & 0.408  & 20342 \\
GPD (12 channels) & - & 72.9 \% & 0.412  & 19756 \\
PointNetGPD \cite{liang2019pointnetgpd} & - & 74.4 \% & 0.434 & 10212 \\
S4g \cite{qin2020s4g} & 0.177 & 83.2 \% & 0.618 & \underline{432}\\
REGNet \cite{zhao2021regnet} & \underline{0.296} & \underline{94.3} \% & \underline{0.662} & 441\\
\hline
HGGD & \textbf{0.503} & \textbf{98.2 \%} & \textbf{0.686}  & \textbf{28} \\
\hline
\end{tabular}
$^1$ Including network inference time and post-processing time.

$^2$ Evaluated with AMD 5600x CPU and single NVIDIA RTX 3060Ti GPU.
\end{threeparttable}
\label{clutter}
\end{center}
\end{table}

\begin{figure}[t]
    \centering
    \includegraphics[width=8 cm]{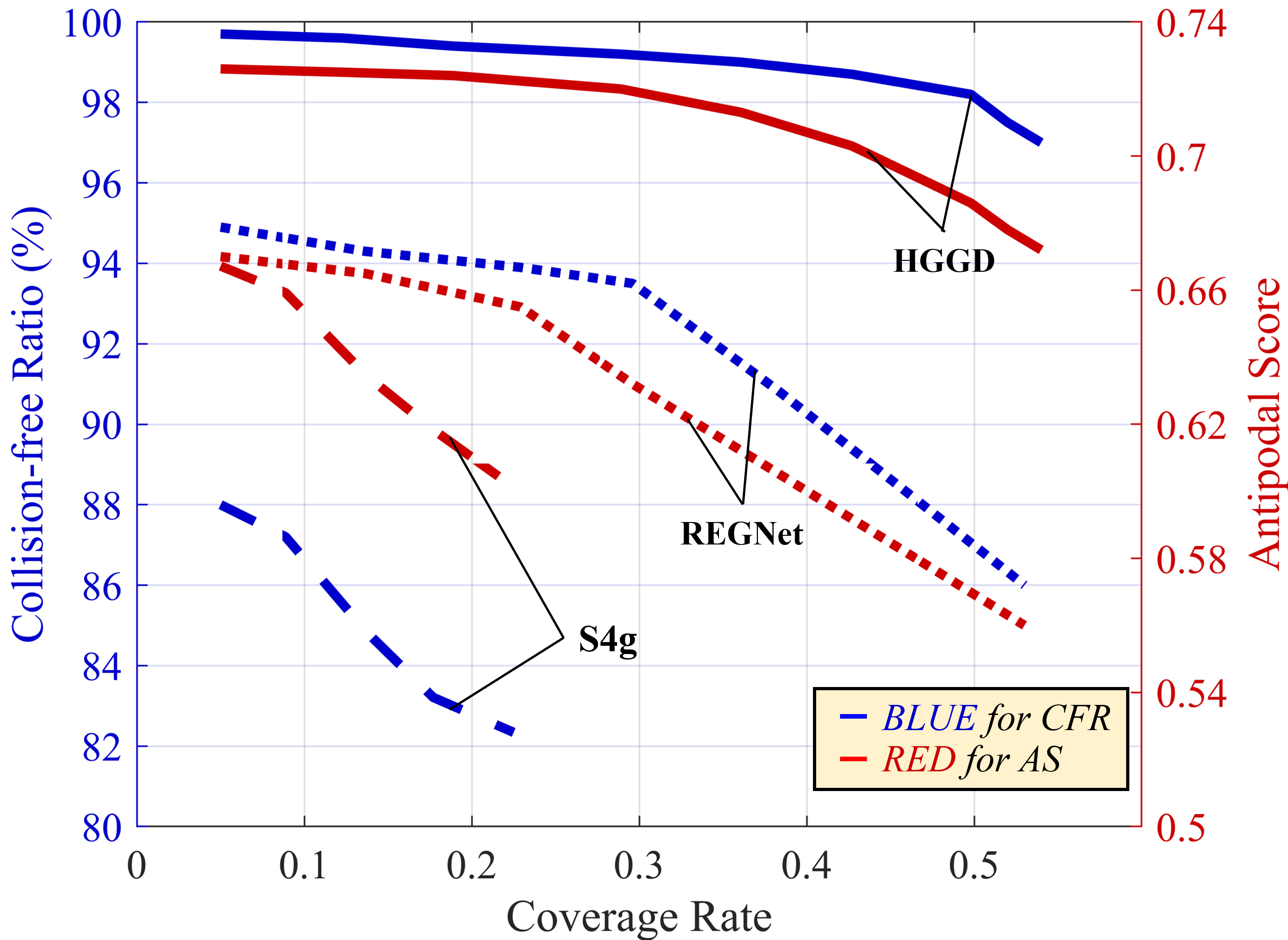}
    \caption{(CFR, CR) and (AS, CR) curves. The red lines represent the Collision-Free Ratio and the blue lines represent the Antipodal Score. When CR increases, compared with baselines, HGGD still remains a relatively high grasp quality.}
    \label{fig:acronym_line}
\end{figure}

\subsection{Evaluation Metrics}

\subsubsection{For TS-ACRONYM}

Following S4g\cite{mousavian20196} and REGNet \cite{zhao2021regnet}, we use \textbf{Collision-Free Ratio (CFR)} and \textbf{Antipodal Score (AS)} to evaluate the quality of each generated grasp. CFR describes the possibility of not colliding with the scene, and AS describes the force closure property. In addition to grasp quality, grasp diversity is important to achieve successful executable grasps. 6-Dof Graspnet \cite{mousavian20196} introduces \textbf{Coverage Rate (CR)}, which describes the diversity of the grasps and measures how well the generated grasps cover all ground truths. Nevertheless, CR defined in \cite{mousavian20196} only considers the distance between grasps, while the grasp rotation is also rather crucial. Thus, according to the rotation distance metrics elaborated in \cite{huynh2009metrics}, in our work, grasp label $\hat{\boldsymbol{g}}$ is regarded as covered if predicted grasp $\boldsymbol{g}$ satisfies the following two conditions:
\begin{itemize}
    \item $\left|\mathbf{t}_{\boldsymbol{g}} - \mathbf{t}_{\hat{\boldsymbol{g}}}\right| \leq 2\ cm$
    \item $1-\left|\mathbf{q}_{\boldsymbol{g}}\cdot \mathbf{q}_{\hat{\boldsymbol{g}}}\right| \leq 0.1$
\end{itemize}
where $\mathbf{t}$ and $\mathbf{q}$ denote translations and quaternions of the grasps. Following \cite{mousavian20196}, we also draw the curves of (CFR, CR) and (AS, CR) by sampling different grasp numbers for detailed analysis and evaluation.

\subsubsection{For GraspNet-1Billion}

GraspNet-1Billion proposes the \textbf{Average Precision (AP)} as an evaluation metric, which calculates the friction coefficient of the top 50 grasp poses by force-closure metric after non-maximum suppression.

\section{EXPERIMENT}

We evaluate our framework both in simulation, including synthetic dataset (TS-ACRONYM), real dataset (GraspNet-1Billion), and in real robot grasping. 

\begin{table*}[t]
\small
\renewcommand{\arraystretch}{1.3}
\caption{Detailed results on GraspNet Dataset, showing APs on RealSense/Kinect split and method time usage}
\begin{center}
\resizebox{\textwidth}{!}{
\begin{threeparttable}
\begin{tabular}{c|c c c|c c c|c c c|c} 
\hline
\multirow{2}{*}{\textbf{Method}}&\multicolumn{3}{c|}{\textbf{Seen}}&\multicolumn{3}{c|}{\textbf{Similar}}&\multicolumn{3}{c|}{\textbf{Novel}}&\textbf{Time$^1$}\\\cline{2-10}
{}&$AP$&$AP_{0.8}$&$AP_{0.4}$&$AP$&$AP_{0.8}$&$AP_{0.4}$&$AP$&$AP_{0.8}$&$AP_{0.4}$&\textbf{/ms}\\
\hline
GPD \cite{ten2017grasp} & 22.87/24.38 & 28.53/30.16 & 12.84/13.46 & 21.33/23.18 & 27.83/28.64 & 9.64/11.32 & 8.24/9.58 & 8.89/10.14 & 2.67/3.16 & -\\
PointnetGPD \cite{liang2019pointnetgpd} & 25.96/27.59 & 33.01/34.21 & 15.37/17.83 & 22.68/24.38 & 29.15/30.84 & 10.76/12.83 & 9.23/10.66 & 9.89/11.24 & 2.74/3.21 & -\\ 
GraspNet-1B \cite{fang2020graspnet} & 27.56/29.88 & 33.43/36.19 & 16.95/19.31 & 26.11/27.84 & 34.18/33.19 & 14.23/16.62 & 10.55/11.51 & 11.25/12.92 & 3.98/3.56 & 296\\
RGB Matters \cite{gou2021rgb} & 27.98/32.08 & 33.47/39.46 & 17.75/20.85 & 27.23/30.40 & 36.34/37.87 & 15.60/18.72 & 12.25/13.08 & 12.45/13.79 & 5.62/6.01 & 440\\
REGNet \cite{zhao2021regnet} & 37.00/37.76 & \nodata/\nodata & \nodata/\nodata & 27.73/28.69 & \nodata/\nodata & \nodata/\nodata & 10.35/10.86 & \nodata/\nodata & \nodata/\nodata & 452 \\
TransGrasp \cite{liu2022transgrasp} & 39.81/35.97 & 47.54/41.69 & 36.42/31.86 & 29.32/29.71 & 34.80/35.67 & 25.19/24.19 & 13.83/11.41 & 17.11/14.42 & 7.67/5.84 & -\\
GSNet \cite{wang2021graspness} & \textbf{67.12}/\textbf{63.50} & \textbf{78.46}/\textbf{74.54} & \underline{60.90}/\textbf{58.11} & \textbf{54.81}/\textbf{49.18} & \textbf{66.72}/\textbf{59.27} & \textbf{46.17}/\textbf{41.89} & \underline{24.31}/\textbf{19.78} & \textbf{30.52}/\textbf{24.60} & \underline{14.23}/\underline{11.17} & $\sim$\underline{100}$^{2}$\\
\hline
HGGD & \underline{64.45}/\underline{61.17} & \underline{72.81}/\underline{69.82} & \textbf{61.16}/\underline{56.52} & \underline{53.59}/\underline{47.02} & \underline{64.12}/\underline{56.78} & \underline{45.91}/\underline{38.86} & \textbf{24.59}/\underline{19.37} & \underline{30.46}/\underline{23.95} & \textbf{15.58}/\textbf{12.14} & \textbf{36}\\
\hline
\end{tabular}
``-'': Result Unavailable.\\
$^1$ Evaluated with AMD 5600x CPU and single NVIDIA RTX 3060Ti GPU.

$^2$ Reported in \cite{wang2021graspness} on NVIDIA RTX 1080Ti GPU since the code is not available.
\end{threeparttable}}
\label{graspnet}
\end{center}
\end{table*}


\subsection{Performance Evaluation}

Firstly, we compare HGGD with state-of-the-arts on TS-ACRONYM, including GPD \cite{ten2017grasp}, PointNetGPD \cite{liang2019pointnetgpd}, S4g \cite{qin2020s4g} and REGNet \cite{zhao2021regnet}. As illustrated in Table \ref{clutter}, HGGD significantly outperforms other methods on CR, AS and CFR metrics, which indicates that HGGD can generate more precise and dense grasps. Meanwhile, Fig. \ref{fig:acronym_line} suggests that HGGD has a more acceptable descending on grasp quality when CR increases. As for time efficiency, owing to its light-weighted architecture and avoiding redundant points processing, HGGD can detect grasps at a real-time speed, {around 28 ms}, much faster than any previous works.

Our model trained on the GraspNet-1Billion dataset is evaluated by AP and compared with other methods. As illustrated in Table \ref{graspnet}, on the premise of ensuring the method's efficiency, our approach achieves nonnegligible AP performance gains on all the seen, similar and novel dataset split compared with REGNet, demonstrating the effectiveness of HGGD. With comparable performance with Graspness \cite{wang2021graspness}, HGGD is nearly three times faster and can run in real-time and performs better in difficult scenarios ($AP_{0.4}$, which means lower friction factor) and unseen scenes.

\begin{figure}[t]
    \centering
    \includegraphics[width=8 cm]{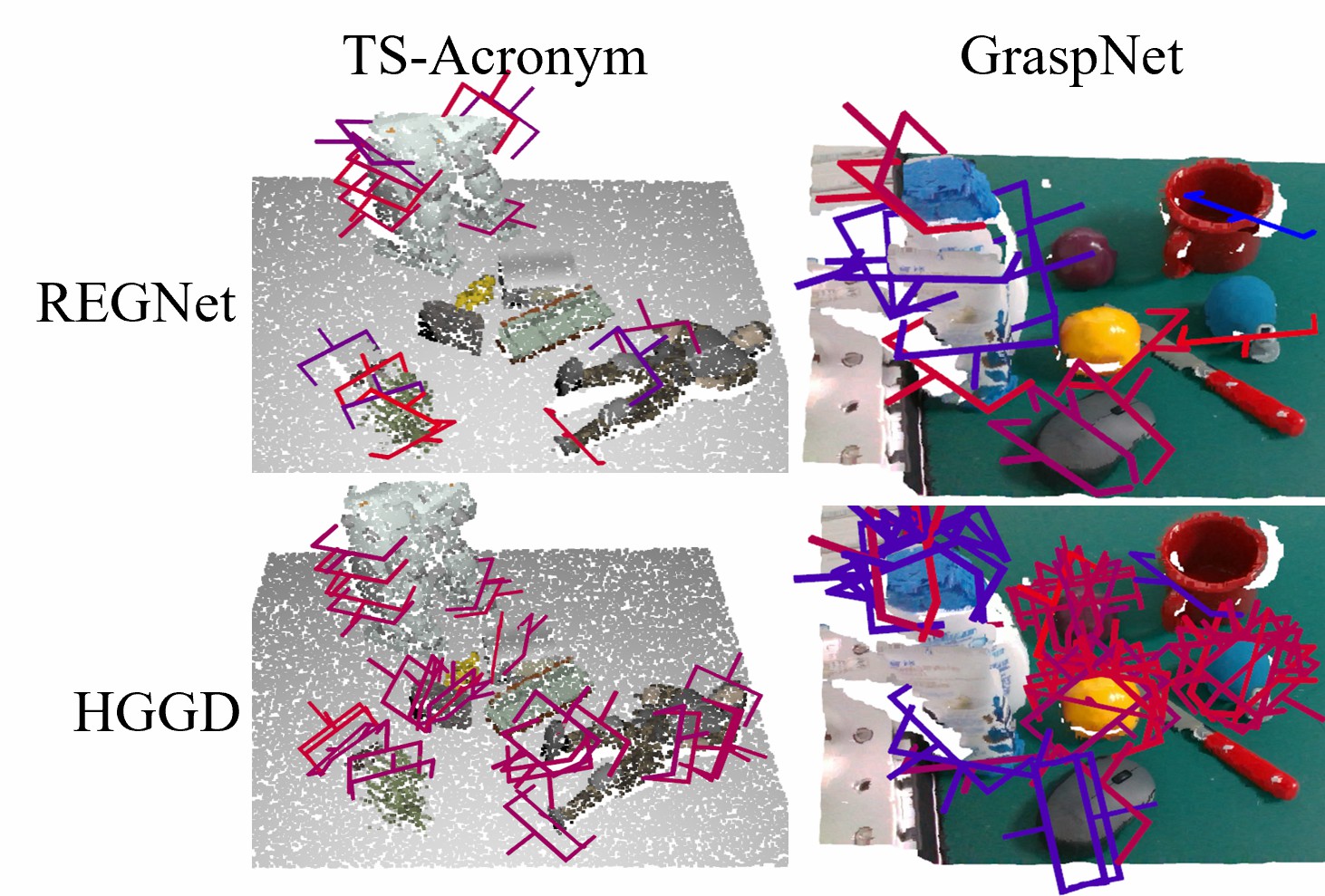}
    \caption{Qualitative results on TS-Acronym and GraspNet-1Billion datasets. Grasps are color-coded based on their test (antipodal/force-closure) scores in RGB space, with red indicating better quality and blue indicating lower quality.}
    \label{fig:compare}
\end{figure}

As is shown in Fig. \ref{fig:compare}, the visualization results also prove that HGGD can predict denser and higher-quality grasps than REGNet, benefiting from the effectiveness of the proposed global-to-local and semantic-to-point scheme.

\subsection{Ablation Studies}



To objectively analyze the role of each module in our method, we build the baseline framework with a random center selection strategy, single-label classification on uniformly sampled anchors and no center refinement for generated grasps. Then we apply the proposed modules to the baseline in order, and experiments show the effectiveness of our method.

\begin{table}[t]
\renewcommand{\arraystretch}{1.25}
\caption{Ablation analysis of each module}
\begin{center}
\begin{threeparttable}
\begin{tabular}{c|c c c c} 
\hline
{TS-ACRONYM}&{\textbf{CR}} $\uparrow$ &{\textbf{CFR}} $\uparrow$ &{\textbf{AS}} $\uparrow$ \\
\hline
baseline & {0.144} & {59.7 \%} & {0.338} & \\
\hline
+ heatmap guidance & 0.450 & 96.9 \% & 0.656 \\
+ center refinement & 0.467 & 97.5 \% & 0.669 \\
+ non-uniform anchor & 0.481 & 97.8 \% & 0.679 \\
+ multi-label classification & 0.498 & \textbf{98.2} \% & \textbf{0.686} \\
+ feature fusion & \textbf{0.503} & \textbf{98.2 \%} & \textbf{0.686} \\
\hline
\end{tabular}
\end{threeparttable}
\label{tab:abl_module}
\end{center}
\end{table}

\begin{table}[t]
\renewcommand{\arraystretch}{1.25}
\caption{Ablation analysis of method robustness}
\begin{center}
\begin{threeparttable}
\begin{tabular}{c|c c c c}
\hline
{TS-ACRONYM with extra noise}&{\textbf{CR}} $\uparrow$ &{\textbf{CFR}} $\uparrow$ &{\textbf{AS}} $\uparrow$ \\
\hline
REGNet & 0.159 & 92.5 \% & 0.629 \\
HGGD w/o feature fusion & 0.464 & 97.5 \% & 0.636 \\
\hline
HGGD & \textbf{0.469} & \textbf{97.9 \%} & \textbf{0.653} \\
\hline
\end{tabular}
\end{threeparttable}
\label{tab:abl_robust}
\end{center}
\end{table}

\begin{table}[t]
\renewcommand{\arraystretch}{1.25}
\caption{Results of robotics experiments}
\begin{center}
\begin{threeparttable}
\begin{tabular}{c|c c c c} 
\hline
\textbf{Scene}&{\textbf{Object}}&{\textbf{Success}}&{\textbf{Attempt}} \\
\hline
1 & 9 & 9 & 10 \\
2 & 8 & 8 & 8 \\
3 & 10 & 10 & 11 \\
4 & 8 & 8 & 9 \\
5 & 9 & 9 & 10 \\
6 & 8 & 8 & 8 \\
7 & 10 & 10 & 10 \\
\hline
\textbf{Success Rate}$^{1}$ & \multicolumn{3}{c}{62 / 66 = \textbf{94\%}} \\
\textbf{Completion Rate}$^{2}$ & \multicolumn{3}{c}{7 / 7 = \textbf{100\%}} \\
\hline
\end{tabular}
$^1$ The sum of \textbf{Attempt} dividing the sum of \textbf{Success}. \\
$^2$ The total scene number dividing the successfully cleared scene number.
\end{threeparttable}
\label{tab:real}
\end{center}
\end{table}

As illustrated in Table \ref{tab:abl_module}, the baseline method shows abysmal performance on each evaluation metric as expected. When the heatmap guidance is offered for local region aggregation, all performance metrics increase significantly, which proves that heatmap guidance is crucial for the grasp detection pipeline to excavate the graspable regions. Ablation experiment for the center refinement module in NMG proves that local geometric features can remarkably increase grasp location precision and reduce the influence of prediction errors in GHM. Then, by adapting anchors during training, HGGD successfully reduces the anchor fitting error and improves detected grasp quality, especially in CR and AS. Beneficial from the non-uniform anchor sampling mechanism, extending traditional single-label classification to the multi-label one is straightforward, enabling HGGD to detect more than one potential graspable rotation in each local region. Multi-label classification successfully improves the grasp quality and diversity of HGGD to a higher level. 

Notably, when the point cloud is unreliable, it is difficult for point-cloud-only methods to mine adequate information for grasp detection. In this circumstance, the semantic information from GHM is helpful due to its containing shape information of objects. To verify this, we apply an extra considerable Gaussian noise to the input point cloud and depth image, significantly reducing the performance of methods without feature fusion, which is quite evident in Table \ref{tab:abl_robust}. Applying feature fusion in NMG can better recover helpful information from the corrupted local point cloud for grasp detection, avoiding more significant performance loss by fully leveraging the semantic and geometrical representations.




\subsection{Training Efficiency}

Benefiting from the proposed global-to-local framework, our method can infer in real-time and be trained efficiently with few key grasp ground truths after non-maximum suppression. Average $\sim500$ grasps per scene ($\sim2\%$ of all labels for GraspNet-1Billion dataset) are sufficient for HGGD to achieve state-of-art performance. Based on its efficient framework and refined dataset organization, HGGD can be fully trained on a single RTX 3090 GPU in 4 hours.

\subsection{Real Robot Experiments}

We also conduct real robot grasping experiments in cluttered scenes on UR-5e with a Robotiq 2-finger parallel-jaw gripper. Realsense-D435i is used to acquire single-view RGBD images. Similar to the experiment procedure of previous works \cite{qin2020s4g,zhao2021regnet,liu2022transgrasp}, we prepare 20 objects of various shapes and sizes used in daily life, then randomly select 8-10 of them for each scene and place them on the table in random poses. The robot performs grasp detection and executes grasping for each scene until no grasp is detected or 15 attempts are tried. We adopt the \textbf{Success Rate} and \textbf{Completion Rate} to evaluate the performance.

Results in Table.\ref{tab:real} reports the performance of HGGD, indicating that HGGD can generalize to the real world to generate high-quality grasps efficiently. Some failures occurred when grasping transparent objects like glass because of the unreliable depth image captured.

\section{CONCLUSION}

This paper presents a novel end-to-end 6-Dof grasp poses detection framework in cluttered scenes. Through the global-to-local and semantic-to-point scheme, HGGD achieves state-of-the-art performance in two representative datasets, much faster than all previous methods. However, our framework is single-view-based and open-loop, inhibiting applications in more complex scenarios. In the future, we intend to utilize it for closed-loop grasp detection, endowing robots to adjust predicted grasp poses while approaching the target object and reacting to the changing environment tactfully.

\bibliography{reference}

\addtolength{\textheight}{-12cm}   

\end{document}